\documentclass[letterpaper, 10 pt, conference]{ieeeconf}  % Comment this line out if you need a4paper

\IEEEoverridecommandlockouts                              % This command is only needed if
\usepackage{graphics} % for pdf, bitmapped graphics files
\usepackage{epsfig} % for postscript graphics files
\usepackage{mathptmx} % assumes new font selection scheme installed
\usepackage{times} % assumes new font selection scheme installed
\usepackage{amsmath} % assumes amsmath package installed
\usepackage{amssymb}  % assumes amsmath package installed
\usepackage{multirow}
\usepackage{subfig}
\usepackage{caption}
\usepackage{color}
\usepackage{url}
\usepackage{verbatim}

\usepackage{cite}
\usepackage[breaklinks,colorlinks,linkcolor=black,citecolor=black,urlcolor=black]{hyperref}

\title{\LARGE \bf
Frustum ConvNet: Sliding Frustums to Aggregate Local Point-Wise Features for Amodal 3D Object Detection
}

\author{Zhixin Wang$^{1}$  and Kui Jia$^{1}$  % <-this % stops a space
\thanks{$^{1}$School of Electronic and Information Engineering, South China University of Technology, Guangzhou, China.
		Email: {\tt\small wang.zhixin@mail.scut.edu.cn}.
		\newline Correspondence to: Kui Jia {\tt\small $<$kuijia@scut.edu.cn$>$}.%
	}
}

\begin{document}
\maketitle
\thispagestyle{empty}
\pagestyle{empty}
%%%%%%%%%%%%%%%%%%%%%%%%%%%%%%%%%%%%%%%%%%%%%%%%%%%%%%%%%%%%%%%%%%%%%%%%%%%%%%%%

\begin{abstract}
	In this work, we propose a novel method termed \emph{Frustum ConvNet (F-ConvNet)} for amodal 3D object detection from point clouds. Given 2D region proposals in an RGB image, our method first generates a sequence of frustums for each region proposal, and uses the obtained frustums to group local points. F-ConvNet aggregates point-wise features as frustum-level feature vectors, and arrays these feature vectors as a feature map for use of its subsequent component of fully convolutional network (FCN), which spatially fuses frustum-level features and supports an end-to-end and continuous estimation of oriented boxes in the 3D space. We also propose component variants of F-ConvNet, including an FCN variant that extracts multi-resolution frustum features, and a refined use of F-ConvNet over a reduced 3D space. Careful ablation studies verify the efficacy of these component variants. F-ConvNet assumes no prior knowledge of the working 3D environment and is thus dataset-agnostic. We present experiments on both the indoor SUN-RGBD and outdoor KITTI datasets. F-ConvNet outperforms all existing methods on SUN-RGBD, and at the time of submission it outperforms all published works on the KITTI benchmark. % We will make the code of F-ConvNet publicly available. 
	Code has been made available at: {\url{https://github.com/zhixinwang/frustum-convnet}.}
\end{abstract}

%%%%%%%%%%%%%%%%%%%%%%%%%%%%%%%%%%%%%%%%%%%%%%%%%%%%%%%%%%%%%%%%%%%%%%%%%%%%%%%%

\section{Introduction}
\label{SecIntro}

Detection of object instances in 3D sensory data has tremendous importance in many applications including autonomous driving, robotic object manipulation, and augmented reality. Among others, RGB-D images and LiDAR point clouds are the most representative formats of 3D sensory data. In practical problems, these data are usually captured by viewing objects/scenes from a single perspective; consequently, only partial surface depth of the observed objects/scenes can be captured. The task of \emph{amodal} 3D object detection is thus to estimate oriented 3D bounding boxes enclosing the full objects, given partial observations of object surface. In this work, we focus on object detection from point clouds, and assume the availability of accompanying RGB images.

Due to the discrete, unordered, and possibly sparse nature of point clouds, detecting object instances from them is challenging and requires learning techniques that are different from the established ones \cite{girshick2015fast, ren2015faster, liu2016ssd} for object detection in RGB images. In order to leverage the expertise in 2D object detection, existing methods convert 3D point clouds either into 2D images by view projection \cite{li2016vehicle, chen2017multi, ku2018joint}, or into regular grids of voxels by quantization\cite{li20173d, wang2015voting, yang2018pixor, liang2018deep}. Although 2D object detection can be readily applied to the converted images or volumes, these methods suffer from loss of critical 3D information in the projection or quantization process.

\begin{figure}
	\begin{center}
	\includegraphics[width=0.8\linewidth]{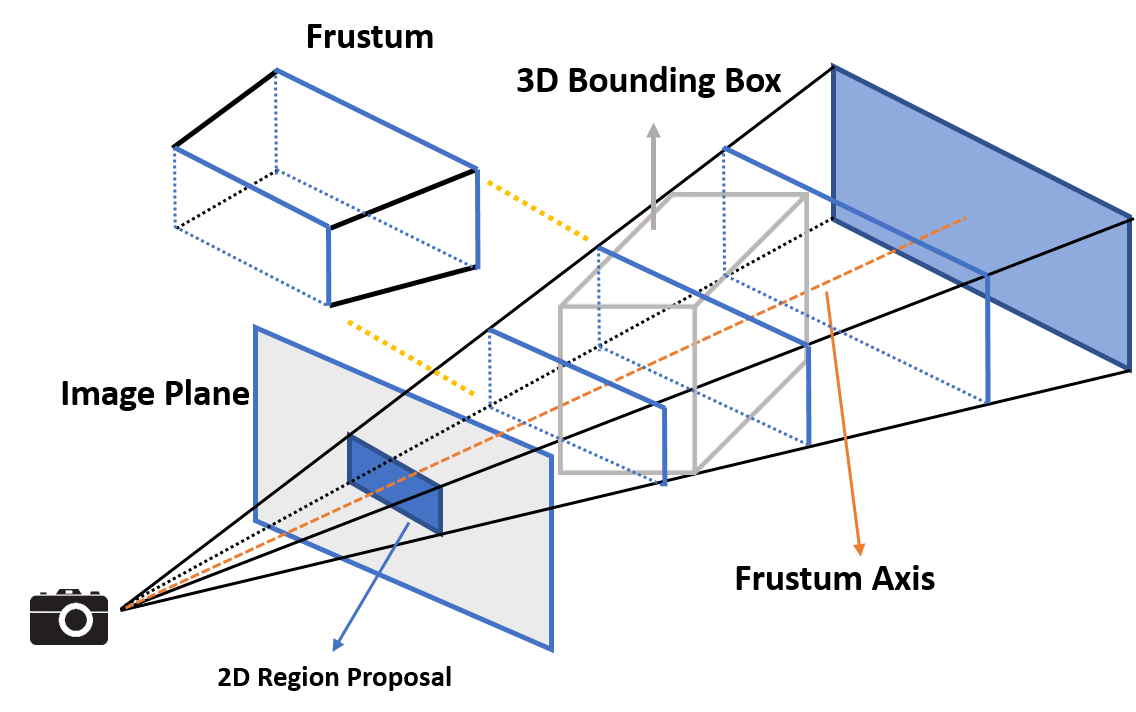}
	\caption[]{Illustration for how a sequence of frustums are generated for a region proposal in an RGB image. }
	\label{Fig:Frustum}
	\end{center}
	\vspace{-0.5cm}
\end{figure}

With the progress of point set deep learning \cite{qi2017pointnet, qi2017pointnet++},  recent methods \cite{qi2018frustum, zhou2018voxelnet} resort to learning features directly from raw point clouds. For example, the seminal work of F-PointNet\cite{qi2018frustum} first finds local points corresponding to pixels inside a 2D region proposal, and then uses PointNet\cite{qi2017pointnet} to segment from these local points the foreground ones; the amodal 3D box is finally estimated from the foreground points. Performance of this method is limited due to the reasons that (1) it is not of end-to-end learning to estimate oriented boxes,  and (2) final estimation relies on too few foreground points which themselves are possibly segmented wrongly.
Methods of VoxelNet style\cite{zhou2018voxelnet, yan2018second, lang2018pointpillars} overcome both of the above limitations by partitioning 3D point cloud into a regular grid of equally spaced voxels; voxel-level features are learned and extracted, again using methods similar to PointNet\cite{qi2017pointnet}, and are arrayed together to form feature maps that are processed subsequently by convolutional (conv) layers; amodal 3D boxes are estimated in an end-to-end fashion using spatially convolved voxel-level features. For the other side of the coin, due to unawareness of objects, sizes and positions of grid partitioning in VoxelNet\cite{zhou2018voxelnet} methods do not take object boundaries into account, and their settings usually assume prior knowledge of the 3D environment (e.g., only one object in vertical space of the KITTI dataset\cite{geiger2012we}), which, however, are not always suitable.

Motivated to address the limitations in\cite{qi2018frustum, zhou2018voxelnet}, we propose in this paper a novel method of amodal 3D object detection termed \emph{Frustum ConvNet (F-ConvNet)}. Similar to \cite{qi2018frustum}, our method assumes the availability of 2D region proposals in RGB images, which can be easily obtained from off-the-shelf object detectors \cite{girshick2015fast, ren2015faster, liu2016ssd}, and identifies 3D points corresponding to pixels inside each region proposal. Different from \cite{qi2018frustum}, our method generates for each region proposal a sequence of (possibly overlapped) \emph{frustums} by sliding along the \emph{frustum axis} \footnote{For any image region, a square pyramid passing though the image region can be specified by the viewing camera and the farthest plane that is perpendicular to the optical axis of the camera. Starting from the image plane, a \emph{frustum} is formed by truncating the pyramid with a pair of parallel planes perpendicular to the optical axis, which is also the \emph{frustum axis}.} (cf. Fig. \ref{Fig:Frustum} for an illustration). These obtained frustums define groups of local points. Given the sequence of frustums and point association, our F-ConvNet starts with lower, parallel layer streams of PointNet style to aggregate point-wise features as a frustum-level feature vector; it then arrays \emph{at its early stage} these feature vectors of individual frustums  as 2D feature maps, and uses a subsequent fully convolutional network (FCN) to down-sample and up-sample frustums such that their features are fully fused across the frustum axis at a higher frustum resolution. Together with a final detection header, our proposed F-ConvNet supports an end-to-end and continuous estimation of oriented 3D boxes, where we also propose an FCN variant that extracts multi-resolution frustum features. Given an initial estimation of 3D box, a final refinement using the same F-ConvNet often improves the performance further. We present careful ablation studies that verify the efficacy of different components of F-ConvNet. On the SUN-RGBD dataset \cite{song2015sun}, our method outperforms all existing ones. On the KITTI benchmark \cite{geiger2012we}, our method outperforms all published works at the time of submission, including those working on point clouds and those working on a combination of point clouds and RGB images. We summarize our contributions as follows.

\begin{itemize}
	\item We propose a novel method termed \emph{Frustum ConvNet (F-ConvNet)} for amodal 3D object detection from point clouds. We use a novel grouping mechanism -- sliding frustums to aggregate local point-wise features for use of a subsequent FCN. Our proposed method supports an end-to-end estimation of oriented boxes in the 3D space that is determined by 2D region proposals.
	% Given a sequence of frustums generated for each region proposal, F-ConvNet aggregates point-wise features as frustum-level feature vectors, and arrays \emph{at its early stage} these feature vectors as a feature map for use of a subsequent FCN, which spatially fuses frustum-level features and supports an end-to-end and continuous estimation of oriented boxes in the 3D space.

	\item We propose component variants of F-ConvNet, including an FCN variant that extracts multi-resolution frustum features, and a refined use of F-ConvNet over a reduced 3D space. Careful ablation studies verify the efficacy of these components and variants.

	\item F-ConvNet assumes no prior knowledge of the working 3D environment, and is thus dataset-agnostic. On the indoor SUN-RGBD dataset \cite{song2015sun}, F-ConvNet outperforms all existing methods; on the outdoor dataset of KITTI benchmark \cite{geiger2012we}, it outperforms all published works at the time of submission.
\end{itemize}

\section{Related Works}
In this section, we briefly review existing methods of amodal 3D object detection. We organize our reviews into two categories of technical approaches, namely those based on conversion of 3D point clouds as images/volumes, and those admitting operation directly on raw point clouds.

\vspace{0.1cm}
\noindent\textbf{Methods based on data conversion} MV3D \cite{chen2017multi} projects LiDAR point clouds to bird eye view (BEV), and then employs a Faster-RCNN \cite{ren2015faster} for 3D object detection. AVOD \cite{ku2018joint} extends MV3D by aggregating the multi-modal features to generate more reliable 3D object proposals. Some existing methods also use depth images as converted data of point clouds. Deng et al. \cite{deng2017amodal} directly estimate 3D bounding boxes from RGB-D images based on the Fast-RCNN framework \cite{girshick2015fast}. Luo et al. \cite{luo2017single} explore the SSD pipeline \cite{liu2016ssd} to fuse RGB and depth images for 3D bounding box estimation. DSS \cite{song2016deep} encodes a depth image  as a grid of 3D voxels by TSDF, and uses 3D CNNs for classification and box estimation.  PIXOR \cite{yang2018pixor} also encodes point clouds as grids of voxels. The above methods based on data conversion can leverage the expertise in mature 2D detection, but the projection or quantization process would cause loss of critical information.

\noindent\textbf{Methods working on raw point clouds} We have reviewed in introduction the seminal work of F-PointNet \cite{qi2018frustum} that works directly on raw point clouds but is not of end-to-end learning to estimate oriented boxes since before that it has to do an instance segmentation and T-Net alignment, and the methods of VoxelNet style \cite{zhou2018voxelnet,  yan2018second, lang2018pointpillars} that resolve this issue but with the shortcoming of object unawareness in 3D point clouds. We note that PointPillars \cite{lang2018pointpillars} explores pillar shape instead of voxel design to aggregate point-wise features. For multi-stage methods \cite{qi2018frustum, yang2018ipod, shi2018pointrcnn}, subsequent works of IPOD \cite{yang2018ipod} and PointRCNN \cite{shi2018pointrcnn} explore different proposal methods. Our proposed F-ConvNet is motivated and designed to combine the benefits of both worlds.

\begin{figure*}
	\begin{center}
	\includegraphics[width=0.85\linewidth]{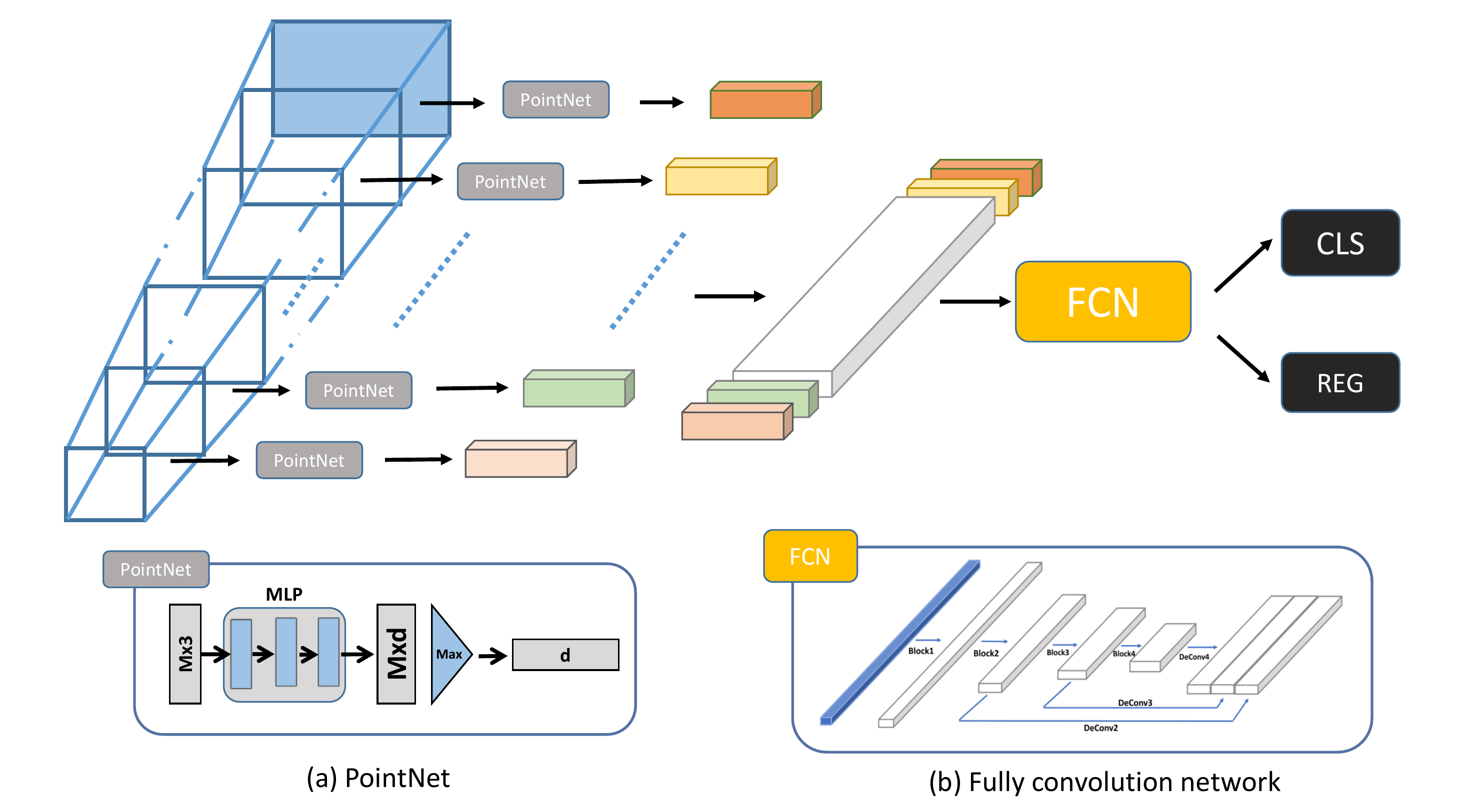}
	\caption[]{The whole framework of our F-ConvNet. We group points and extract features by PointNet from a sequence of frustums, and for 3D box estimation, frustum-level features are re-formed as a 2D feature map for use of our fully convolutional network (FCN) and detection header (CLS and REG). (a) The architecture of PointNet. (b) The architecture of FCN used in Frustum ConvNet for KITTI dataset. Each convolutional layer is followed by Batch Normalization and ReLU nonlinearity. Blue-colored bar in (b) represents the 2D feature map of arrayed frustum-level feature vectors. }
	\label{Fig:Framework}
	\end{center}
	\vspace{-0.5cm}
\end{figure*}
\section{The Proposed Frustum ConvNet}

In this section, we present our proposed Frustum ConvNet (F-ConvNet) that supports end-to-end learning of amodal 3D object detection. Design of F-ConvNet centers on the notion of \emph{square frustum}, and a sequence of frustums along the same \emph{frustum axis} connect a cloud of discrete, unordered points with an FCN that enables oriented 3D box estimation in a continuous 3D space. Fig.\ref{Fig:Framework} give an illustration. By assuming the availability of 2D region proposals in RGB images, we will first introduce our way of point association with sequences of (possibly overlapped) frustums that are obtained by sliding along frustum axes determined by 2D region proposals, and compare with alternative ways of point association/grouping. We will then present the architecture of F-ConvNet, and specify how point-wise features inside individual frustums are aggregated and re-formed as 2D feature maps for a continuous frustum-level feature fusion and 3D box estimation. We finally explain how an F-ConvNet can be trained using losses borrowed from the literature of 2D object detection.

\subsection{Associating Point Clouds with Sliding Frustums}

Learning semantics from point clouds grounds on extraction of low-level geometric features that are defined over local groups of neighboring points. Due to the discrete, unordered nature of point cloud, there exists no oracle way to associate individual points into local groups. In the literature of point set classification/segmentation \cite{qi2017pointnet++, li2018pointcnn}, local groupings are formed by searching nearest neighbors, with seed points sampled from a point cloud using farthest point sampling (FPS). FPS can efficiently cover a point cloud, but it is unaware of object positions; consequently, grouping based on FPS is not readily useful for tasks concerning with detection of object instances from a point cloud. Rather than sampling seed points for local groupings, methods of VoxelNet style \cite{zhou2018voxelnet, yan2018second, lang2018pointpillars} define a regular grid of equally spaced voxels in the 3D space, and points falling in same voxels are grouped together. Although voxels may densely cover the entire 3D space, their sizes and grid positions do not take object boundaries into account. In addition, their settings usually assume prior knowledge of the 3D environment and the contained object categories (e.g., car and pedestrian in the KITTI dataset \cite{geiger2012we}), which, however, are not always available.

\begin{figure}
	\begin{center}
	\includegraphics[width=0.8\linewidth]{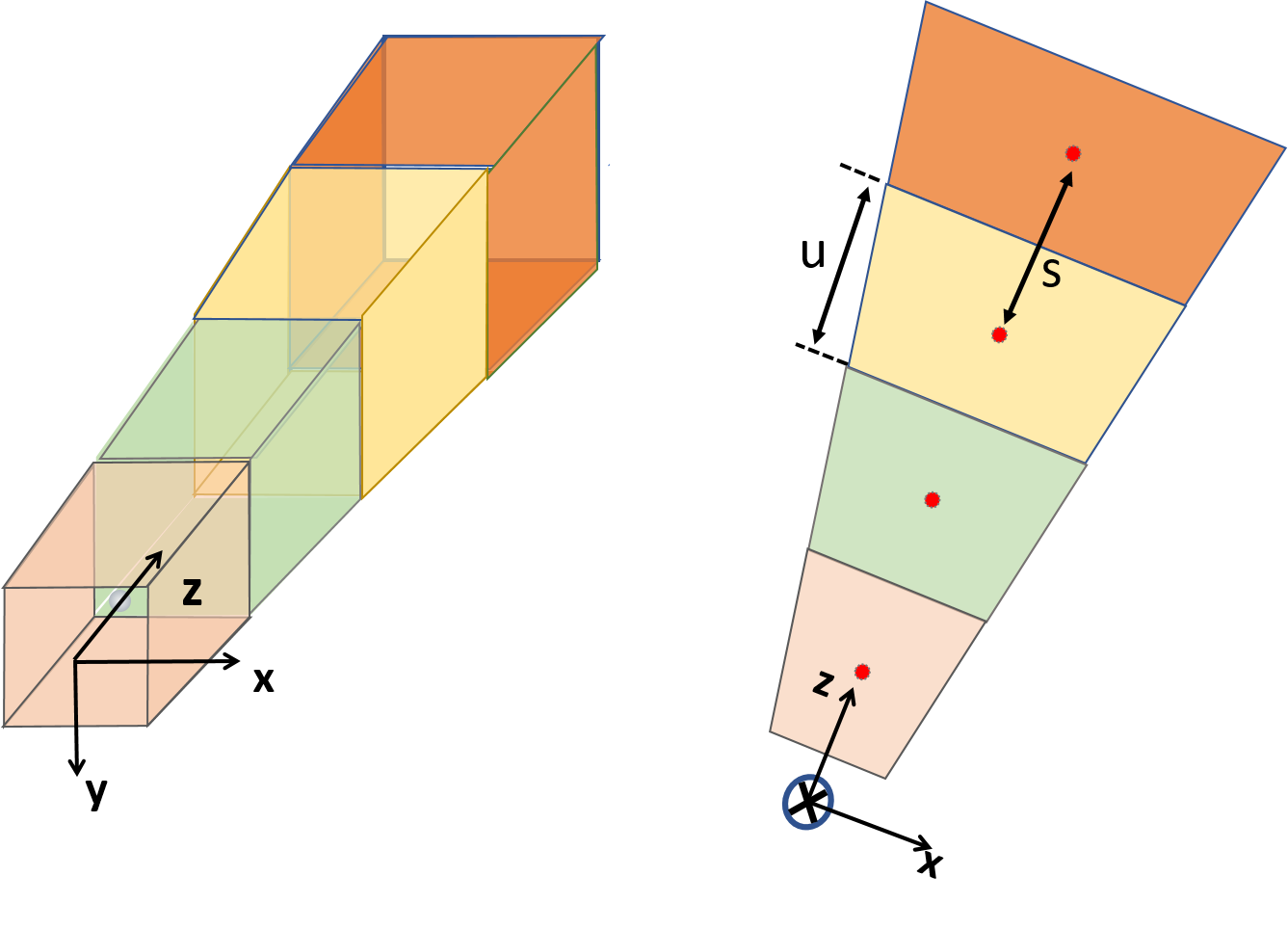}
	\caption[]{An illustration of our frustums (coded as different colors) for point association and frustum-level feature aggregation. A sequence of non-overlapped frustums are shown here for simplicity. Actually, we set $u=2s$ in our experiments. We show the top view on the right to denote clearly the sliding stride $s$ and height $u$ of frustums.  }
	\label{Fig:SFGP}
	\end{center}
	\vspace{-0.5cm}
\end{figure}

To address these limitations, we propose the following scheme to group local points. We assume that an RGB image is available accompanying the 3D point cloud, and that 2D region proposals are also provided by off-the-shelf object detectors \cite{girshick2015fast, ren2015faster, liu2016ssd}. A sequence of (possibly overlapped) frustums can be obtained by sliding a pair of parallel planes along the frustum axis with an equal stride, where the pair of planes are also perpendicular to the frustum axis. We also assume the optical axis of the camera is perpendicular to this 2D region, which suggests an initial adjustment of camera coordinate system has already been performed, as shown in Fig.\ref{Fig:SFGP}. We generate such a sequence of frustums for each 2D region proposal, and we use thus obtained frustum sequences to group points, i.e., points falling inside the same frustums are grouped together. Assuming that 2D region proposals are accurate enough, our frustums mainly contain foreground points, and are aware of object boundaries. We note that for each 2D region proposal, a single frustum of larger size (defined by the image plane and the farthest plane) is generated in \cite{qi2018frustum} and all points falling inside this frustum are grouped together; consequently, an initial stage of foreground point segmentation has to be performed before amodal 3D box estimation. In contrast, we generate for each region proposal a sequence of frustums whose feature vectors are arrayed as a feature map and used in a subsequent FCN for an end-to-end estimation of oriented boxes in the continuous 3D space, as presented shortly.

\subsection{The Architecture of Frustum ConvNet}

Given a sequence of frustums generated from a region proposal, the key design of an F-ConvNet is to aggregate \emph{at its early stage} point-wise features inside each frustum as a frustum-level feature vector, and then array as a 2D feature map these feature vectors of individual frustums for use of a subsequent FCN, which, together with a detection header, supports an end-to-end and continuous estimation of oriented 3D boxes. Fig. \ref{Fig:Framework} gives the architecture. We present separate components of the F-ConvNet as follows.

\vspace{0.1cm}
\noindent\textbf{Frustum-level feature extraction via PointNet}
For a 2D region proposal in an RGB image, assume a sequence of $T$ frustums of height $u$ are generated by sliding along the frustum axis with a stride $s$. For any one of them, assume it contains $M$ local points whose coordinates in the camera coordinate system are denoted as $\{ \mathbf{x}_i = (x_i, y_i, z_i) \}_{i=1}^{M}$. To learn and extract point-wise features, we use PointNet\cite{qi2017pointnet} in this work that stacks three fully-connected (FC) layers, followed by a final layer that aggregates features of individual points as a frustum-level feature vector via element-wise max pooling, as shown in Fig.\ref{Fig:Framework}(a).  We apply the PointNet to each frustum, and thus the $T$ duplicate PointNets, with shared weights, form the lower, parallel streams of our F-ConvNet. Instead of using $\{ \mathbf{x}_i \}_{i=1}^{M}$ as input of PointNet directly, we use relative coordinates $\{ \bar{\mathbf{x}}_i = (\bar{x}_i, \bar{y}_i, \bar{z}_i) \}_{i=1}^{M}$ that are obtained by subtracting each $\mathbf{x}_i$ with the centroid $\mathbf{c}$ of the frustum, i.e., $\bar{\mathbf{x}}_i = \mathbf{x}_i - \mathbf{c}$ for $i = 1, \dots, M$. We note that choices other than PointNet are applicable as well, such as PointCNN \cite{li2018pointcnn}.

% We apply a PointNet to each frustum, and the $T$ duplicate PointNets, with shared weights, form the lower, parallel streams of our F-ConvNet.  Our used PointNet consists of three FC layers which respectively output features of dimensions $d/2$, $d/2$, and $d$.

%\begin{figure}
%	\centering
%	\includegraphics[width=0.8\linewidth]{figs2/FCN_wo_header}
%	\caption[]{The architecture of fully convolutional network (FCN) in Frustum ConvNet. Each convolutional layer is followed by Batch Normalization and ReLU nonlinearity. Blue color represents 2D feature map of arrayed frustum-level feature vectors. }
%	\label{Fig:FCN}
%\end{figure}

\vspace{0.1cm}
\noindent\textbf{Fully convolutional network}
Denote the extracted frustum-level feature vectors as $\{ \mathbf{f}_i \}_{i=1}^L$, with $\mathbf{f}_i \in \mathbb{R}^d$. We array these $L$ vectors to form a 2D feature map $\mathbf{F}$ of the size $L\times d$, which will be used as input of a subsequent FCN. As shown in Fig.\ref{Fig:Framework}(b), our FCN consists of blocks of conv layers, and de-conv layers corresponding to each block. Convolution in conv layers is applied across the frustum dimension by using kernels of the size $3\times d$. The final layer of each of the conv blocks, except the first block, also down-samples (halves) the 2D feature map at the frustum dimension by using stride-$2$ convolution. Convolution and down-sampling fuse features across frustums and produce at different conv blocks \emph{virtual frustums} of varying heights (along the frustum axis direction). Given output feature map of each conv block, a corresponding de-conv layer is used that up-samples at the frustum dimension the feature map to a specified (higher) resolution $\tilde{L}$; outputs of all de-conv layers are then concatenated together along the feature dimension. % It outputs a feature map of the size $\tilde{L}\times 3d$.
Feature concatenation from virtual frustums of varying sizes provides a hierarchical granularity of frustum covering, which would be useful to estimate 3D boxes of object instances whose sizes are unknown and vary. In this work, we use an FCN of $4$ conv blocks and $3$ de-conv layers for KITTI, and an FCN of $5$ conv blocks and $4$ de-conv layers for SUN-RGBD. Layer specifics of these FCNs are given in the appendix.

\begin{figure}
	\begin{center}
	\includegraphics[width=0.8\linewidth]{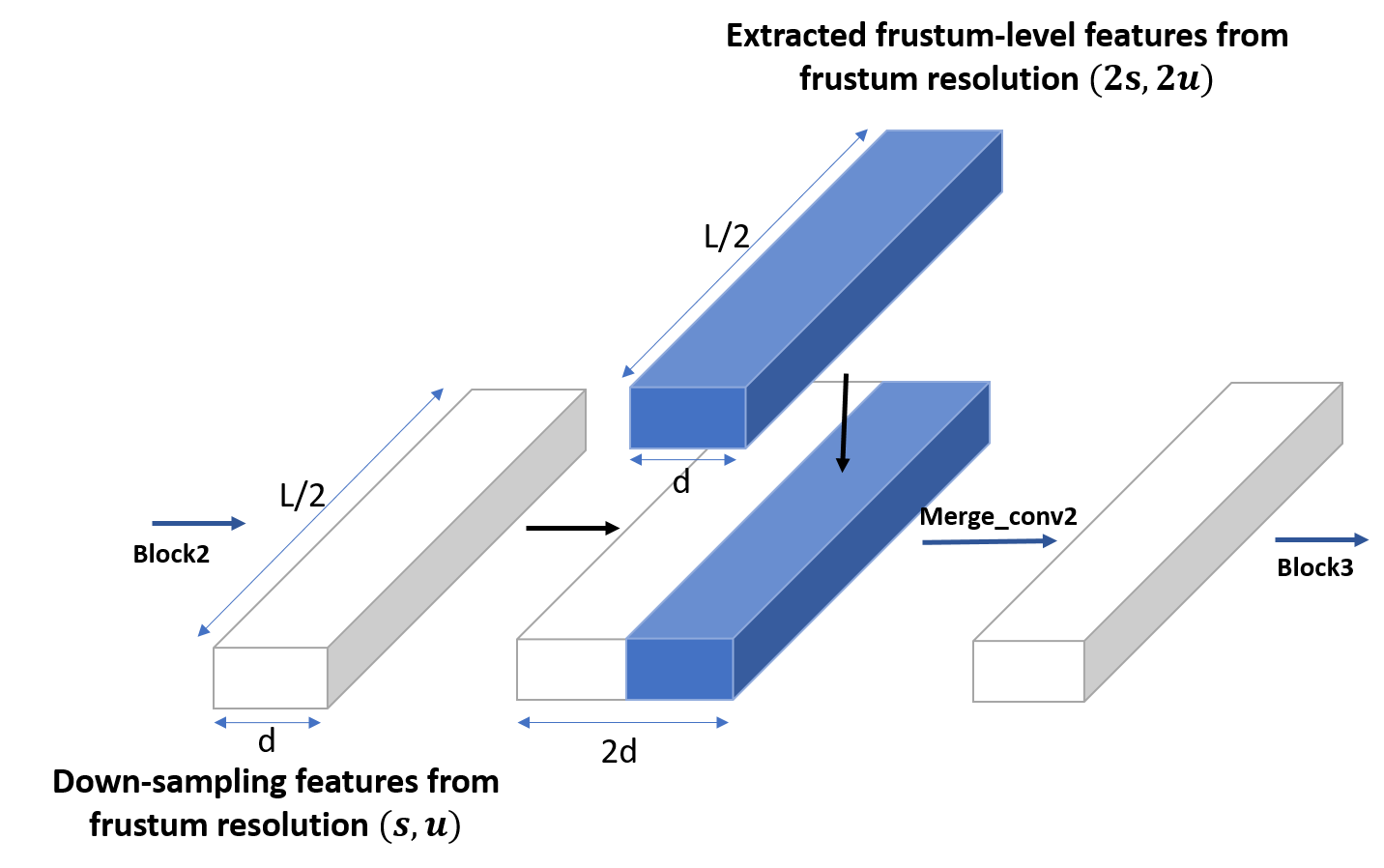}
	\caption[]{Illustration of our multi-resolution frustum feature integration. We show an example between Block2 and Block3. We also use it in between Block3 and Block4, and between Block4 and DeConv4. Including the first resolution, we have in total four kinds of resolutions in KITTI dataset.}
	\label{Fig:MSF}
	\end{center}
	\vspace{-0.5cm}
\end{figure}

\vspace{0.1cm}
\noindent\textbf{A multi-resolution frustum feature integration variant}
We already know that output feature maps of conv blocks in FCN are of reduced resolutions by a power of $2$ at the frustum dimension. Take the one with the size of $L/2 \times d$ as an example. For the same 2D region proposal, a new sequence of $T/2$ frustums can be generated by sliding along the frustum axis with a stride $2s$. Applying PointNet to each of the generated frustums and arraying the resulting feature vectors produce a new feature map of the same size $L/2 \times d$. When frustum height is doubled as $2u$, the new sequence covers the same 3D space at a half coarser resolution, while its feature map being compatible with its corresponding one in FCN. We then concatenate along the feature dimension the two feature maps of the same size, giving rise to a new one of the size $L/2 \times 2d$. A final conv layer is used to resize it back as a feature map of size $L/2 \times d$, so that it can be placed back in FCN with no change of other FCN layers. Fig.\ref{Fig:MSF} illustrates the above procedure. The procedure can be used for each down-sampled feature maps in FCN. We refer to this scheme as a multi-resolution frustum feature integration variant of F-ConvNet. Ablation studies in Section \ref{Sec:Multi-Resolution} verify its efficacy.

\subsection{Detection Header and Training of Frustum ConvNet}

On top of FCN is the detection header composed of two, parallel conv layers, as shown in Fig.\ref{Fig:Framework}. They are respectively used as the classification and regression branches. The whole F-ConvNet is trained using a multi-task fashion, similar to those in 2D object detection \cite{girshick2015fast,ren2015faster, liu2016ssd}.

Suppose we have $K$ object categories. The classification branch is trained to output a $\tilde{L}\times (K+1)$ frustum-wise probability map of object categories, plus the background one. In this work, we use focal loss \cite{lin2017focal} for classification branch to cope with imbalance of foreground and background samples. 

Ground truth of an oriented 3D bounding box is parameterized as $\{x_c^g, y_c^g, z_c^g, l^g, w^g, h^g, \theta^g\}$, where $\{x_c^g, y_c^g, z_c^g\}$ denote coordinates of box center,  $\{l^g, w^g, h^g\}$ denote three side lengths of the box, and $\theta^g$ denotes the yaw angle that means the in-plane rotation perpendicular to the gravity direction. We discretize the range [$-\pi$, $\pi$) of yaw angles into $N$ bins, and define for each frustum $KN$ anchor boxes, i.e., $N$ ones per foreground category. For any one of them parameterized as $\{x_c^a, y_c^a, z_c^a, l^a, w^a, h^a, \theta^a\}$, we use centroid of the frustum as $\{x_c^a, y_c^a, z_c^a\}$, compute from training samples the category-wise averages of the three side lengths as $\{l^a, w^a, h^a\}$, and set $\theta^a$ as one of the bin centers of yaw angles. This gives the following offset formulas:
\begin{equation}
	\begin{split}
		& \Delta x = x_c^g - x_c^a, \Delta y = y_c^g - y_c^a, \Delta z = z_c^g - z_c^a, \\
		& \Delta l = \frac{l^g - l^a}{l_a}, \Delta w = \frac{w^g - w^a}{w_a}, \Delta h = \frac{h^g - h^a}{h^a}, \\
		& \Delta \theta = \theta^g - \theta^a .
	\end{split}
\end{equation}

Regression loss for the center is based on the Euclidean distance and smooth $l1$ regression loss for offsets of size and angle. Besides, we also use a corner loss \cite{qi2018frustum} to regularize box regression of all parameters. Together with the focal loss for classification branch, the whole F-ConvNet is trained using a total of three losses.

\subsection{Final Refinement}

\begin{figure}
	\begin{center}
	\includegraphics[width=0.9\linewidth]{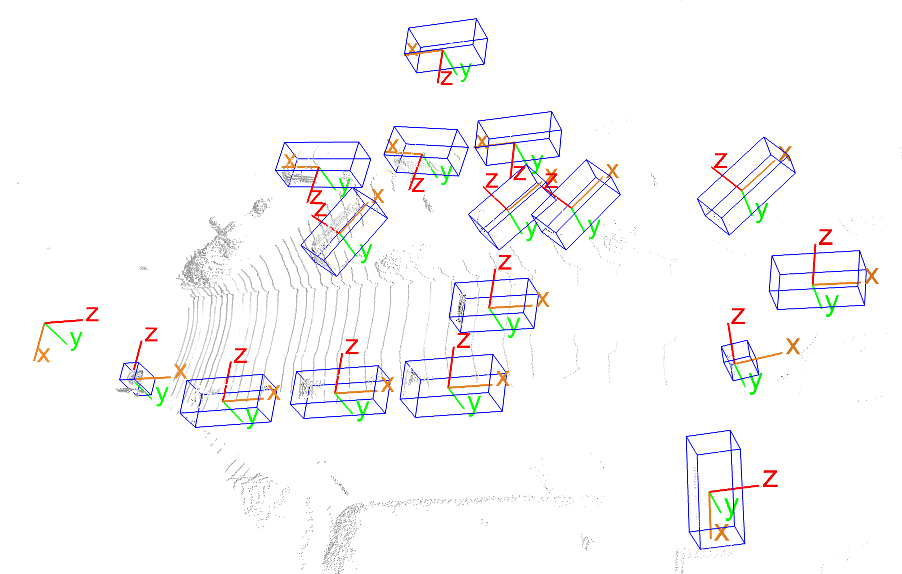}
	\caption[]{Normalization of point coordinates for final refinement. Leftmost legend is the camera coordinate system. Normalized coordinate systems are shown with each cuboid box, where the origin of each system is located at each cuboid center and its direction is aligned with our first predicted box. }
	\label{Fig:coordnorm}
	\end{center}
	\vspace{-0.5cm}
\end{figure}

We have assumed for now that the 2D region proposals are accurate enough. In practice, region proposals provided by object detectors do not bound object instances precisely. As a remedy, we propose a refinement that applies the same F-ConvNet architecture to points falling inside the oriented 3D boxes that we have just estimated. Specifically, we first expand each estimated box by a specified factor --- we set the factor as $1.2$ in this work, and normalize points inside the expanded box by translation and rotation, as shown in Fig. \ref{Fig:coordnorm}. These normalized points are used as input of a second F-ConvNet for a final refinement. Ablation studies show that this refinement resolves well the issue caused by inaccurate 2D object proposals.

\section{Experiments}
\subsection{Datasets and Implementation Details}

\noindent\textbf{KITTI} The KITTI dataset \cite{geiger2012we} contains 7,481 training pairs and 7,518 testing pairs of RGB images and point clouds of three object categories (i.e., Car, Pedestrian, and Cyclist). For each category, detection results are evaluated based on three levels of difficulty (i.e., easy, moderate, and hard). We train two separate F-ConvNets respectively for Car and Pedestrian/Cyclist. Since ground truth of the test set is unavailable, we follow existing works \cite{chen2017multi} and split the original training set into the new training and validation ones respectively of 3,712 and 3,769 samples. We conduct our ablation experiments using this splitting, and our final results on the KITTI test set are obtained by server submission. We use the official 3D IoU evaluation metrics of 0.7, 0.5, and 0.5 respectively for the categories of Car, Cyclist, and Pedestrian.

\vspace{0.1cm}
\noindent\textbf{SUN-RGBD} The SUN-RGBD dataset \cite{song2015sun} contains 10,355 RGB-D images (5,285 training ones and 5,050 testing ones) of 10 object categories. We convert the depth images as point clouds for use of our method. Results are evaluated on the 10 categories under 0.25 3D IoU threshold. For this dataset, we do not use the final refinement of our method.

\vspace{0.1cm}
\noindent\textbf{Implementation details} Our use of 2D object detectors is as follows. For the KITTI validation dataset, we use the 2d detection results provided by F-PointNet \cite{qi2018frustum}. For the KITTI test dataset, we use RRC \cite{ren2017accurate} for the car category and MSCNN \cite{cai2016unified} for the pedestrians and cyclist categories. We directly use the release models provided by these methods. As for SUN-RGBD, we train Faster-RCNN \cite{ren2015faster} with the backbone network of ResNet-50 \cite{he2016deep}. We do data augmentation to the obtained 2D region proposals by translation and scaling during training. We randomly sample from 3D points corresponding to each region proposal to have a fixed number 1,024 for KITTI and 2,048 for SUN-RGBD. For final refinement, we use a fixed number of 512. We also do random flipping and shifting to these points, similar to \cite{qi2018frustum}.

To prepare positive and negative training samples, we shrink ground-truth boxes by a ratio of $0.5$, and count anchor boxes whose centers fall in the shrunken ground-truth boxes as foreground ones, count the others as background. We ignore the anchor boxes whose centers fall between the shrunken boxes and ground-truth boxes.  We train F-ConvNets with a mini-batch size 32 on one GPU. We use ADAM optimizer with weight decay of  $0.0001$. Learning rates start from 0.001 and decay by a factor of 10 every $20^{th}$ epoch of the total 50 epoches. We consider a depth range of $[0, 70]$ meters in KITTI and that of $[0, 8]$ in SUN-RGBD. For KITTI, we use $4$ frustum resolutions of $u = [0.5, 1.0, 2.0, 4.0]$ and $s = [0.25, 0.5, 1.0, 2.0]$ for the car category, with $d=[128, 128, 256, 512]$, $L=280$, and $\tilde{L}=140$, and $4$ frustum resolutions of $u = [0.2, 0.4, 0.8, 1.6]$ and $s = [0.1, 0.2, 0.4, 0.8]$ for the pedestrian and cyclist categories, with $d=[128, 128, 256, 512]$, $L=700$, and $\tilde{L}=350$. For SUN-RGBD, we use $5$ frustum resolutions of $u = [0.2, 0.4, 0.8, 1.6, 3.2]$ and $s = [0.1, 0.2, 0.4, 0.8, 1.6]$, with $d=[128, 128, 256, 512, 512]$, $L=80$, and $\tilde{L}=40$.

At evaluation time, we only keep predicted foreground samples and apply an NMS module with a 3D IoU threshold of $0.1$ to reduce redundancy.
The final 3D detection scores are computed by adding 2D detection scores and predicted 3D bounding box scores.

\subsection{Ablation Studies}

In this section, we verify components and variants of our proposed F-ConvNet by conducting ablation studies on the train/val split of KITTI. We follow the convention and use the car category that contains the most training examples. Before individual studies, we first report in Tab.\ref{Tab:KITTI_VAL_3D} and Tab.\ref{Tab:KITTI_VAL_BEV} our results of 3D detection and BEV detection on the validation set.
% Our method outperforms existing ones on both of the two tasks.
For the most important ``Moderate'' column, our method outperforms existing ones on both of the two tasks.
\begin{table}
	\begin{center}
		\begin{tabular}{c|ccc}
			\hline
			Method                            & Easy           & Moderate       & Hard           \\ \hline
			MV3D\cite{chen2017multi}          & 71.29          & 62.68          & 56.56          \\
			VoxelNet\cite{zhou2018voxelnet}   & 81.97          & 65.46          & 62.85          \\
			F-PointNet\cite{qi2018frustum}    & 83.76          & 70.92          & 63.65          \\
			AVOD-FPN\cite{ku2018joint}        & 84.41          & 74.44          & 68.65          \\
			ContFusion\cite{liang2018deep}    & 86.32          & 73.25          & 67.81          \\
			IPOD \cite{yang2018ipod}          & 84.1           & 76.4           & 75.3           \\
			PointRCNN \cite{shi2018pointrcnn} & 88.88          & 78.63          & \textbf{77.38} \\ \hline
			Ours                              & \textbf{89.02} & \textbf{78.80} & 77.09          \\ \hline
		\end{tabular}
		\caption{3D object detection AP (\%) on KITTI val set.}
		\label{Tab:KITTI_VAL_3D}
	\end{center}
	\vspace{-0.5cm}
\end{table}
\begin{table}
	\begin{center}
		\begin{tabular}{c|ccc}
			\hline
			Method                          & Easy           & Moderate       & Hard           \\ \hline
			MV3D\cite{chen2017multi}        & 86.55          & 78.10          & 76.67          \\
			VoxelNet\cite{zhou2018voxelnet} & 89.60          & 84.81          & 78.57          \\
			F-PointNet\cite{qi2018frustum}  & 88.16          & 84.92          & 76.44          \\
			ContFusion\cite{liang2018deep}  & \textbf{95.44} & 87.34          & 82.43          \\
			IPOD \cite{yang2018ipod}        & 88.3           & 86.4           & 84.6           \\ \hline
			Ours                            & 90.23          & \textbf{88.79} & \textbf{86.84} \\ \hline
		\end{tabular}
		\caption{BEV detection AP (\%) on KITTI val set.}
		\label{Tab:KITTI_VAL_BEV}
	\end{center}
\vspace{-0.5cm}
\end{table}

\vspace{0.1cm}
\noindent\textbf{Influence of 2D region proposal} Our method relies on accuracy of 2D region proposals. To investigate how much it affects the performance, we use three 2D object detectors with increased practical performance, namely a baseline Faster-RCNN \cite{ren2015faster} with a backbone network of ResNet-50 \cite{he2016deep}, a detector provided by \cite{qi2018frustum}, and an oracle one of ground-truth 2D boxes. Results in Tab.\ref{Tab:2DDectors} confirm that better performance of 2D detection positively affects our method.

\begin{table}[h]
	\begin{center}
	\begin{tabular}{ccc|ccc}
		\hline
		\multicolumn{3}{c|}{2D Detection} & \multicolumn{3}{c}{3D Detection}                                     \\ \hline
		Easy                              & Moderate                         & Hard   & Easy  & Moderate & Hard  \\ \hline
		96.19                             & 87.51                            & 77.41  & 85.19 & 74.05    & 64.77 \\
		96.48                             & 90.31                            & 87.63  & 86.51 & 76.57    & 68.17 \\
		100.00                            & 100.00                           & 100.00 & 87.68 & 85.47    & 78.19 \\ \hline
	\end{tabular}
	\caption{Influence of 2D region proposal. Each line corresponds to results from a different 2D object detector. }
	\label{Tab:2DDectors}
	\end{center}
	\vspace{-0.5cm}
\end{table}

\vspace{0.1cm}
\noindent\textbf{Effect of frustum feature extractor} We use PointNet \cite{qi2017pointnet} to extract and aggregate point-wise features as frustum-level feature vectors. Other choices such as PointCNN \cite{li2018pointcnn} are applicable as well. To compare, we replace the element-wise max pooling in PointNet by the X-Conv operation in PointCNN for feature aggregation. Tab.\ref{Tab:XConv} shows that PointCNN is also a possible choice of frustum feature extractor; however, its performance is not necessarily better than the simple PointNet. We use one resolution for this experiment.

\begin{table}[ht]
	\begin{center}
		\begin{tabular}{c|ccc}
			\hline
			         & Easy  & Moderate & Hard  \\ \hline
			PointNet & 84.09 & 75.32    & 67.45 \\
			PointCNN & 81.91 & 73.83    & 66.37 \\ \hline
		\end{tabular}
		\caption{Comparison between frustum feature extractors.}
		\label{Tab:XConv}
	\end{center}
	\vspace{-0.5cm}
\end{table}

\vspace{0.1cm}
\label{Sec:Multi-Resolution}
\noindent\textbf{Effect of the multi-resolution frustum feature integration variant} To investigate the effect of this variant, we plug in various resolution combinations into F-ConvNet. Results in Tab.\ref{Tab:MultiResolution} confirm the efficacy.

%\begin{table}[ht]
%    \centering
%    \begin{tabular}{cccc|ccc}
%        \hline
%        $(0.25, 0.5)$  & $(0.5, 1.0)$   &$(1.0, 2.0)$   &$(2.0, 4.0)$   & Easy  & Moderate & Hard  \\ \hline
%        \checkmark &            &            &            & 84.09 & 75.32    & 67.45 \\
%        \checkmark & \checkmark &            &            & 84.19 & 74.88    & 66.95 \\
%        \checkmark &            & \checkmark &            & 85.41 & 75.63    & 67.44 \\
%        \checkmark &            &            & \checkmark & 86.12 & 76.04    & 67.97 \\
%        \checkmark & \checkmark & \checkmark &            & 86.21 & 76.12    & 67.96 \\
%        \checkmark &            & \checkmark & \checkmark & 86.69 & 76.30    & 68.02 \\
%        \checkmark & \checkmark & \checkmark & \checkmark & 86.51 & 76.57    & 68.17 \\
%        \hline
%    \end{tabular}
%    \caption{Investigation of the multi-resolution frustum feature integration variant. }
%    \label{Tab:MultiResolution}
%\end{table}

\begin{table}[ht]
	\begin{center}
	\begin{tabular}{ccc|ccc}
		\hline
		$(0.5, 1.0)$ & $(1.0, 2.0)$ & $(2.0, 4.0)$ & Easy  & Moderate & Hard  \\ \hline
		             &              &              & 84.09 & 75.32    & 67.45 \\
		\checkmark   &              &              & 84.19 & 74.88    & 66.95 \\
		             & \checkmark   &              & 85.41 & 75.63    & 67.44 \\
		             &              & \checkmark   & 86.12 & 76.04    & 67.97 \\
		\checkmark   & \checkmark   &              & 86.21 & 76.12    & 67.96 \\
		             & \checkmark   & \checkmark   & 86.69 & 76.30    & 68.02 \\
		\checkmark   & \checkmark   & \checkmark   & 86.51 & 76.57    & 68.17 \\
		\hline
	\end{tabular}
	\caption{Investigation of the multi-resolution frustum feature integration variant. We show different combinations of pair $(s, u)$, where $s$ denotes sliding stride of frustums and $u$ for height of frustums. }
	\label{Tab:MultiResolution}
	\end{center}
	\vspace{-0.5cm}
\end{table}

% \vspace{0.1cm}
\noindent\textbf{Effect of focal loss and final refinement} We use focal loss\cite{lin2017focal} to cope with imbalance of foreground and background training samples. We also propose a final refinement step to cope with less accurate 2D region proposals. The effects of these two components are clearly demonstrated in Tab.\ref{Tab:FocalLoss}.

\begin{table}[ht]
	\begin{center}
		\begin{tabular}{c|ccc}
			\hline
			                  & Easy  & Moderate & Hard  \\ \hline
			w/o FL and w/o RF & 83.78 & 74.05    & 65.96 \\
			w/o RF            & 86.51 & 76.57    & 68.17 \\
			Ours              & 89.02 & 78.80    & 77.09 \\ \hline
		\end{tabular}
		\caption{Effects of focal loss (FL) and final refinement (RF).}
		\label{Tab:FocalLoss}	
	\end{center}
	\vspace{-0.5cm}
\end{table}

\begin{table*}[h]
	\begin{center}
		\begin{tabular}{c|ccc|ccc|ccc}
			\hline
			\multirow{2}{*}{Method}         & \multicolumn{3}{c|}{Cars} & \multicolumn{3}{c|}{Pedestrians} & \multicolumn{3}{c}{Cyclists}                                                                                                       \\ \cline{2-10}
			                                & Easy                      & Moderate                         & Hard                         & Easy           & Moderate       & Hard           & Easy           & Moderate       & Hard           \\ \hline
			MV3D\cite{chen2017multi}        & 71.09                     & 62.35                            & 55.12                        & -              & -              & -              & -              & -              & -              \\
			VoxelNet\cite{zhou2018voxelnet} & 77.47                     & 65.11                            & 57.73                        & 39.48          & 33.69          & 31.51          & 61.22          & 48.36          & 44.37          \\
			F-PointNet\cite{qi2018frustum}  & 81.20                     & 70.29                            & 62.19                        & 51.21          & 44.89          & 40.23          & 71.96          & 56.77          & 50.39          \\
			AVOD-FPN\cite{ku2018joint}      & 81.94                     & 71.88                            & 66.38                        & 50.80          & 42.81          & 40.88          & 64.00          & 52.18          & 46.61          \\
			SECOND\cite{yan2018second}      & 83.13                     & 73.66                            & 66.20                        & 51.07          & 42.56          & 37.29          & 70.51          & 53.85          & 46.90          \\
			IPOD\cite{yang2018ipod}         & 79.75                     & 72.57                            & 66.33                        & \textbf{56.92} & 44.68          & \textbf{42.39} & 71.40          & 53.46          & 48.34          \\
			PointPillars\cite{lang2018pointpillars}
			                                & 79.05                     & 74.99                            & 68.30                        & 52.08          & 43.53          & 41.49          & 75.78          & 59.07          & 52.92          \\
			PointRCNN\cite{shi2018pointrcnn}
			                                & \textbf{85.94}            & 75.76                            & \textbf{68.32}               & 49.43          & 41.78          & 38.63          & 73.93          & 59.60          & 53.59          \\ \hline
			Ours                            & 85.88                     & \textbf{76.51}                   & 68.08                        & 52.37          & \textbf{45.61} & 41.49          & \textbf{79.58} & \textbf{64.68} & \textbf{57.03} \\ \hline
		\end{tabular}
		\caption{3D object detection AP (\%) on KITTI test set.}
		\label{Tab:KITTI_TEST_3D}
	\end{center}
	\vspace{-0.5cm}
\end{table*}

\begin{table*}[h]
	\begin{center}
		\begin{tabular}{c|ccc|ccc|ccc}
			\hline
			\multirow{2}{*}{Method}         & \multicolumn{3}{c|}{Cars} & \multicolumn{3}{c|}{Pedestrians} & \multicolumn{3}{c}{Cyclists}                                                                                                       \\ \cline{2-10}
			                                & Easy                      & Moderate                         & Hard                         & Easy           & Moderate       & Hard           & Easy           & Moderate       & Hard           \\ \hline
			MV3D\cite{chen2017multi}        & 86.02                     & 76.90                            & 68.49                        & -              & -              & -              & -              & -              & -              \\
			VoxelNet\cite{zhou2018voxelnet} & 77.47                     & 65.11                            & 57.73                        & 39.48          & 33.69          & 31.51          & 61.22          & 48.36          & 44.37          \\
			F-PointNet\cite{qi2018frustum}  & 88.70                     & 84.00                            & 75.33                        & 58.09          & 50.22          & \textbf{47.57} & 75.38          & 61.96          & 54.68          \\
			AVOD-FPN\cite{ku2018joint}      & 88.53                     & 83.79                            & 77.90                        & 58.75          & 51.05          & 47.54          & 68.06          & 57.48          & 50.77          \\
			SECOND\cite{yan2018second}      & 88.07                     & 79.37                            & 77.95                        & 55.10          & 46.27          & 44.76          & 73.67          & 56.04          & 48.78          \\
			IPOD\cite{yang2018ipod}         & 86.93                     & 83.98                            & 77.85                        & \textbf{60.83} & \textbf{51.24} & 45.40          & 77.10          & 58.92          & 51.01          \\
			PointPillars\cite{lang2018pointpillars}
			                                & 88.35                     & \textbf{86.10}                   & \textbf{79.83}               & 58.66          & 50.23          & 47.19          & 79.14          & 62.25          & 56.00          \\
			PointRCNN\cite{shi2018pointrcnn}
			                                & 89.47                     & 85.68                            & 79.10                        & 55.92          & 47.53          & 44.67          & 81.52          & 66.77          & \textbf{60.78} \\ \hline
			Ours                            & \textbf{89.69}            & 83.08                            & 74.56                        & 58.90          & 50.48          & 46.72          & \textbf{82.59} & \textbf{68.62} & 60.62          \\ \hline
		\end{tabular}
		\caption{3D object localization AP (BEV) (\%) on KITTI test set. }
		\label{Tab:KITTI_TEST_BEV}
	\end{center}
	\vspace{-0.5cm}
\end{table*}

\begin{figure*}[h]
	\centering
	\includegraphics[width=1.0\linewidth]{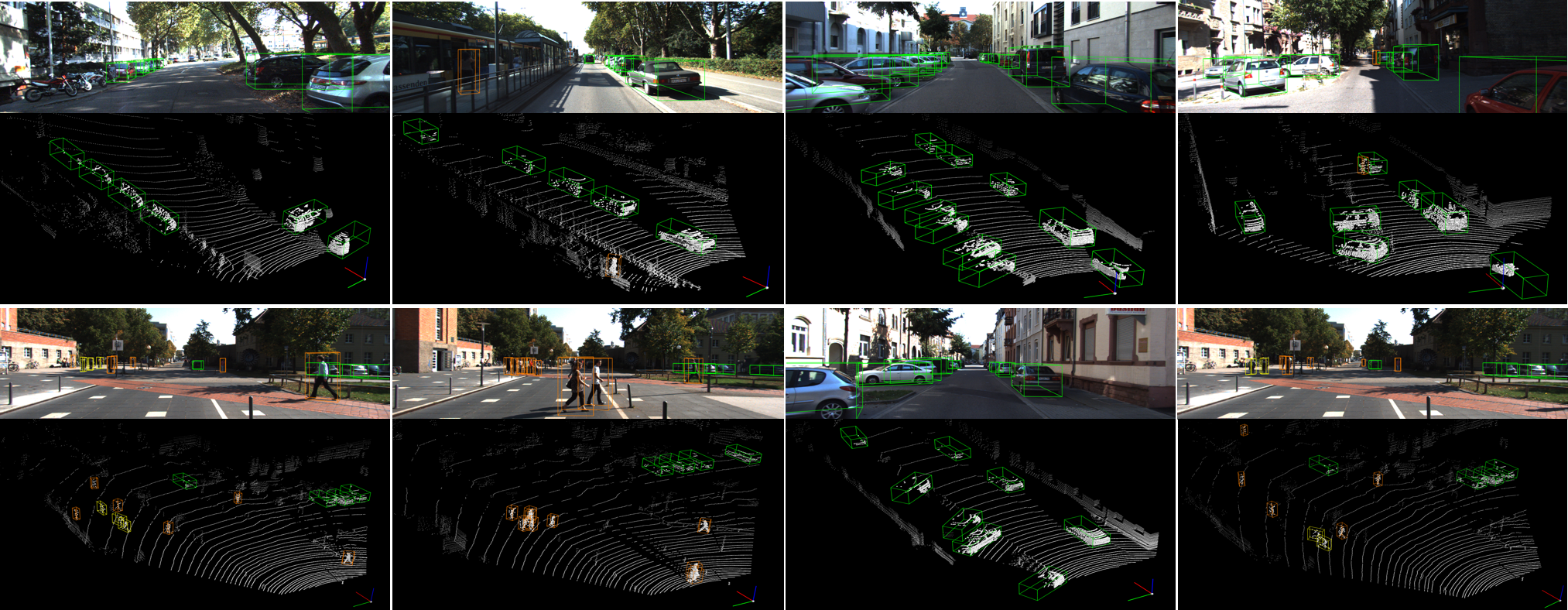}
	\caption[]{Qualitative results on the KITTI test set. Best view in color with zoom-in.
		Different color bounding boxes denote different categories, with green for car, orange for pedestrian, and yellow for cyclist. }
	\label{Fig:DEMOKITTI}
	\vspace{-0.3cm}
\end{figure*}

\begin{table*}[h]
	\begin{center}
		\begin{tabular}{c|cccccccccc|c}
			\hline
			\multicolumn{1}{c|}{Method}         & bathtub       & bed            & bookshelf      & chair          & desk           & dresser        & nightstand     & soft           & table         & toilet        & mean           \\ \hline
			DSS\cite{song2016deep}              & 44.2          & 78.8           & 11.9           & 61.2           & 20.5           & 6.4            & 15.4           & 53.5           & 50.3          & 78.9          & 42.1           \\
			COG\cite{ren2016three}              & 58.26         & 63.67          & 31.80          & 62.17          & \textbf{45.19} & 15.47          & 27.36          & 51.02          & 51.29         & 70.07         & 47.63          \\
			2Ddriven3D\cite{lahoud20172d}       & 43.45         & 64.48          & 31.40          & 48.27          & 27.93          & 25.92          & 41.92          & 50.39          & 37.02         & 80.40         & 45.12          \\
			PointFusion\cite{xu2018pointfusion} & 37.26         & 68.57          & \textbf{37.69} & 55.09          & 17.16          & 23.95          & 32.33          & 53.83          & 31.03         & 83.80         & 45.38          \\
			Ren et al.\cite{ren20183d}          & \textbf{76.2} & 73.2           & 32.9           & 60.5           & 34.5           & 13.5           & 30.4           & 60.4           & \textbf{55.4} & 73.7          & 51.0           \\
			F-PointNet\cite{qi2018frustum}      & 43.3          & 81.1           & 33.3           & 64.2           & 24.7           & 32.0           & 58.1           & 61.1           & 51.1          & \textbf{90.9} & 54.0           \\ \hline
			Ours                                & 61.32         & \textbf{83.19} & 36.46          & \textbf{64.40} & 29.67          & \textbf{35.10} & \textbf{58.42} & \textbf{66.61} & 53.34         & 86.99         & \textbf{57.55} \\ \hline
		\end{tabular}
		\caption{3D object detection AP (\%) on the SUN-RGBD test set (IoU 0.25). }
		\label{Tab:SUNRGBD_TEST}
		\vspace{-0.5cm}
	\end{center}

\end{table*}

\subsection{Comparisons with the State of the Art}

\noindent\textbf{The KITTI Results} Tab.\ref{Tab:KITTI_TEST_3D} shows the performance of our method on the KITTI test set, which is obtained by server submission. Our method outperforms all existing published works, and at the time of submission it ranks $4^{th}$ on the KITTI leaderboard.  We also show performance of our method on 3D object localization in Tab. \ref{Tab:KITTI_TEST_BEV}. For this detection task, the 3D bounding boxes are projected to bird-eye view plane and IoU is evaluated on oriented 2D boxes. Representative results of our method are visualized in Fig.\ref{Fig:DEMOKITTI}.

\vspace{0.1cm}
\noindent\textbf{The SUN-RGBD Results} We also apply our proposed F-ConvNet to the indoor environment of SUN-RGBD. Our results in Tab.\ref{Tab:SUNRGBD_TEST} are better than those of all existing methods, showing the general usefulness of our proposed method.

\section{Conclusion}

We have presented a novel method of Frustum ConvNet (F-ConvNet) for amodal 3D object detection in an end-to-end and continuous fashion. The proposed method is dataset-agnostic and demonstrates state-of-the-art performance on both the indoor SUN-RGBD and outdoor KITTI datasets. The method is useful for many applications such as autonomous driving and robotic object manipulation. In future research, we will investigate more seamless ways of integrating point-wise and RGB features and we expect even better performance would be achieved.


\section*{Appendix}

Layer specifics of the FCN component of F-ConvNet for the KITTI and SUN-RGBD datasets are respectively given in Tab.\ref{Tab:KITII_SPECIFICS} and Tab.\ref{Tab:SUNRGBD_SPECIFICS}.

\begin{table}[h]
	\begin{center}
	\begin{tabular}{c|c}
		\hline
		Name                    & Kernel size/Filter no./Striding/Padding \\ \hline
		Block1                  & 3$\times$128 / 128 / 1 / 1              \\
		\hline
		\multirow{2}{*}{Block2} & 3$\times$128 / 128 / 2 / 1              \\
		                        & 3$\times$128 / 128 / 1 / 1              \\
		\hline
		\multirow{2}{*}{Block3} & 3$\times$128 / 256 / 2 / 1              \\
		                        & 3$\times$256 / 256 / 1 / 1              \\
		\hline
		\multirow{2}{*}{Block4} & 3$\times$256 / 512 / 2 / 1              \\
		                        & 3$\times$512 / 512 / 1 / 1              \\
		\hline
		Deconv2                 & 1$\times$128 / 256 / 1 / 0              \\
		Deconv3                 & 2$\times$256 / 256 / 2 / 0              \\
		Deconv4                 & 4$\times$512 / 256 / 4 / 0              \\ \hline
		Merge\_conv2            & 1$\times$256 / 128 / 1 / 0              \\
		Merge\_conv3            & 1$\times$512 / 256 / 1 / 0              \\
		Merge\_conv4            & 1$\times$1024 / 512 / 1 / 0             \\
		\hline
	\end{tabular}
	\caption{Layer specifics of the FCN component of F-ConvNet for KITTI. }
	\label{Tab:KITII_SPECIFICS}
	\end{center}
	\vspace{-0.5cm}
\end{table}

\begin{table}[h]
	\begin{center}
	\begin{tabular}{c|c}
		\hline
		Name                    & Kernel size/Filter no./Striding/Padding \\ \hline
		Block1                  & 3$\times$64 / 64  / 1 / 1               \\
		\hline
		\multirow{2}{*}{Block2} & 3$\times$64 / 128  / 2 / 1              \\
		                        & 3$\times$128 / 128 / 1 / 1              \\
		\hline
		\multirow{2}{*}{Block3} & 3$\times$128 / 256 / 2 / 1              \\
		                        & 3$\times$256 / 256 / 1 / 1              \\
		\hline
		\multirow{2}{*}{Block4} & 3$\times$256 / 512 / 2 / 1              \\
		                        & 3$\times$512 / 512 / 1 / 1              \\
		\hline
		\multirow{2}{*}{Block5} & 3$\times$512 / 512 / 2 / 1              \\
		                        & 3$\times$512 / 512 / 1 / 1              \\
		\hline
		Deconv2                 & 1$\times$128 / 256 / 1 / 0              \\
		Deconv3                 & 2$\times$256 / 256 / 2 / 0              \\
		Deconv4                 & 4$\times$512 / 256 / 4 / 0              \\
		Deconv5                 & 8$\times$512 / 256 / 8 / 0              \\
		\hline
		Merge\_conv2            & 1$\times$256 / 128 / 1 / 0              \\
		Merge\_conv3            & 1$\times$512 / 256 / 1 / 0              \\
		Merge\_conv4            & 1$\times$1024 / 512 / 1 / 0             \\
		Merge\_conv5            & 1$\times$1024 / 512 / 1 / 0             \\
		\hline
	\end{tabular}
	\caption{Layer specifics of the FCN component of F-ConvNet for SUN-RGBD. }
	\label{Tab:SUNRGBD_SPECIFICS}
	\end{center}
	\vspace{-0.5cm}
\end{table}


\section*{ACKNOWLEDGMENT}
This work is supported in part by the National Natural Science Foundation of China (Grant No.: 61771201), and the Program for Guangdong Introducing Innovative and Enterpreneurial Teams (Grant No.: 2017ZT07X183).

\bibliographystyle{IEEEtran}
\bibliography{egbib}

%%%%%%%%%%%%%%%%%%%%%%%%%%%%%%%%%%%%%%%%%%%%%%%%%%%%%%%%%%%%%%%%%%%%%%%%%%%%%%%%
%%%%%%%%%%%%%%%%%%%%%%%%%%%%%%%%%%%%%%%%%%%%%%%%%%%%%%%%%%%%%%%%%%%%%%%%%%%%%%%%
% References are important to the reader; therefore, each citation must be complete and correct. If at all possible, references should be commonly available publications.


%%%%%%%%%%%%%%%%%%%%%%%%%%%%%%%%%%%%%%%%%%%%%%%%%%%%%%%%%%%%%%%%%%%%%%%%%%%%%%%%
\end{document}